\documentclass{article} % For LaTeX2e
\usepackage{gpl_iclr2022_conference,times}

\usepackage{amsmath,amsfonts,bm}

\def\Figref#1{Figure~\ref{#1}}

\def\Secref#1{Section~\ref{#1}}
\def\eqref#1{equation~\ref{#1}}
\def\1{\bm{1}}

\DeclareMathAlphabet{\mathsfit}{\encodingdefault}{\sfdefault}{m}{sl}
\SetMathAlphabet{\mathsfit}{bold}{\encodingdefault}{\sfdefault}{bx}{n}

\usepackage{url}
\usepackage{hyperref}
\usepackage{cleveref} % always load after hyperref (never before)
\usepackage{booktabs}
\usepackage{authblk}

\title{Towards Flexible Inference in \\Sequential Decision Problems \\via Bidirectional Transformers}

\author[1]{\textbf{Micah Carroll}}%\thanks{Part of this work was done during an internship at Microsoft.}
\author[1]{\textbf{Jessy Lin}}
\author[1]{\textbf{Orr Paradise}}
\author[2]{\textbf{Raluca Georgescu}}
\author[2]{\textbf{Mingfei Sun}}
\author[2]{\textbf{David Bignell}}
\author[2]{\textbf{Stephanie Milani}}
\author[2]{\textbf{Katja Hofmann}}
\author[2]{\textbf{Matthew Hausknecht}}
\author[1]{\textbf{Anca Dragan}}
\author[2]{\textbf{Sam Devlin}}
\affil[1]{UC Berkeley}
\affil[2]{Microsoft Research}

\usepackage{xspace}
\usepackage{wrapfig}
\usepackage{graphicx}
\usepackage{ifthen}
\newboolean{include-notes}
\setboolean{include-notes}{false}

\DeclareRobustCommand{\MicahComment}[1]{\ifthenelse{\boolean{include-notes}}
 {{\color{cyan}M: #1}}{}}
\DeclareRobustCommand{\JessyComment}[1]{\ifthenelse{\boolean{include-notes}}
 {{\color{orange}J: #1}}{}}
\DeclareRobustCommand{\sm}[1]{\ifthenelse{\boolean{include-notes}}
 {{\color{pink}S: #1}}{}}
\DeclareRobustCommand{\orrp}[1]{\ifthenelse{\boolean{include-notes}}
 {{\color{blue}O: #1}}{}}
\DeclareRobustCommand{\orrpfn}[1]{\ifthenelse{\boolean{include-notes}}
 {{\footnote{\color{blue}O: #1}}}{}}
\DeclareRobustCommand{\adnote}[1]{\ifthenelse{\boolean{include-notes}}
 {{\color{red}A: #1}}{}}
\DeclareRobustCommand{\todo}[1]{\ifthenelse{\boolean{include-notes}}
 {{\color{red} #1}}{}}

\newcommand{\shortblank}{\rule{0.25cm}{0.15mm}\xspace}
\newcommand{\longblank}{\rule{0.5cm}{0.15mm}\xspace}
\newcommand{\masking}{masking\xspace}
\newcommand{\maskings}{maskings\xspace}

\newcommand{\fb}{\textit{Flexi}\textnormal{BiT}\xspace} % Need \tetxtnormal so it doesn't render differently in, say, headings
\newcommand{\Prob}[1]{\mathbb{P}(#1)}
\newcommand{\prg}[1]{\noindent\textbf{#1}}

\iclrfinalcopy % Uncomment for camera-ready version, but NOT for submission.
\begin{document}

\maketitle

\textbf{Note: This is paper is superseded by the full version \citep{main}.}

\begin{abstract}
Randomly masking and predicting word tokens has been a successful approach in pre-training language models for a variety of downstream tasks. In this work, we observe that the same idea also applies naturally to sequential decision making, where many well-studied tasks like behavior cloning, offline RL, inverse dynamics, and waypoint conditioning correspond to different sequence maskings over a sequence of states, actions, and returns. We introduce the \fb{} framework, which provides a unified way to specify models which can be trained on many different sequential decision making tasks. We show that \textit{a single \fb model} is simultaneously capable of carrying out many tasks with performance similar to or better than specialized models. Additionally, we show that performance can be further improved by fine-tuning our general model on specific tasks of interest.
\end{abstract}

\section{Introduction}

% \MicahComment{tasks vs models vs inferences}
% \MicahComment{masking schemes / input maskings}

% \orrp{Figure out \masking vs. masking vs. masking schemes etc.}

\label{sec:intro}
Masked language modeling~\citep{devlin2018bert} is a key technique in natural language processing (NLP). Under this paradigm, models are trained to predict randomly-masked subsets of tokens in a sequence. For example, during training, a BERT model might be asked to predict the missing words in the sentence ``yesterday I \longblank{} cooking a \longblank''. Importantly, while unidirectional models like GPT~\citep{radford2018improving} are trained to predict the next token conditioned only on the left context (making their most natural usage that of language \textit{generation}), bidirectional models trained on this objective learn to model both the left \emph{and} right context to represent each word token. This leads to richer representations that can then be fine-tuned to excel on a variety of downstream tasks~\citep{devlin2018bert}.

Our work investigates how masked modeling can be a powerful idea in sequential decision problems. Consider a sequence of states $s$ and actions $a$ collected across $T$ timesteps $s_1,a_1,\dots,s_T,a_T$. If we consider each state and action as tokens of a sequence (analogous to words in NLP) and mask the last action, say $(s_1,a_1,s_2,a_2,s_3,\shortblank)$, predicting the missing token $a_3$ amounts to a Behavior Cloning prediction with two timesteps of history \citep{pomerleau_efficient_1991}, given that this masking corresponds to the inference $\Prob{a_3 | s_{1:3}, a_{1:2}}$. From this perspective, training a model to predict missing tokens from all \maskings of the form $(s_1,a_1,\dots,s_{t}, \shortblank, \dots, \shortblank)$ 
for all $t \in [1, \dots, T]$ 
corresponds to training a Behavior Cloning (BC) model.

Similarly, many well-studied inference tasks can be expressed as a \masking: goal or waypoint conditioned BC~\citep{ding_goal-conditioned_2020, rhinehart_precog_2019}, offline reinforcement learning (RL) \citep{levine2020offline}, forward or inverse dynamics prediction \citep{ha_world_2018, christiano_transfer_2016, chen_transdreamer_2022}, initial-state inference \citep{shah_preferences_2019}, and more.
With this idea, we introduce the \fb framework: \textbf{Flexi}ble Inference in Sequential Decision Problems via \textbf{Bi}directional \textbf{T}ransformers. In this framework, a model can be trained to perform a variety of inference tasks in a unified way by expressing each inference of interest as a masking scheme and training on all such masking schemes. In contrast to standard approaches that train a specialized model for each inference task, we show how a single \fb{} model can be trained to perform a large variety of tasks out-of-the-box (even without fine-tuning).

We test this framework in a gridworld navigation task. We train a \fb model using a random masking scheme, which corresponds to training a single model to perform all possible inference tasks by sampling among them. We show how this scheme enables a single \fb{} model to condition on arbitrary subsets of states, actions, and rewards to perform a variety of useful inference tasks. We then systematically analyze how the masking schemes seen at training time affect downstream task performance. We find that, compared to specialized models that only train on the task of interest, a \fb model trained on random masking performs better than many of them without fine-tuning. Fine-tuning such a generic \fb model on any task of interest further improves performance. 

% We supplement these findings with preliminary experiments in continuous-control robotics environments, which further support our claims. 
Our results suggest that expressing tasks as sequence maskings with the \fb{} framework may be a promising unifying approach to building general-purpose ``multi-task'' models, or simply offer an avenue for building better-performing single-task models via unified multi-task training.

\section{Related Work}

%The successes of modern natural language processing (NLP) techniques have motivated efforts to apply similar ingredients -- unsupervised pre-training and Transformers~\cite{vaswani2017attention} -- to reinforcement learning (RL). Recent work has shown that GPT-style~\cite{radford2018improving} Transformer models can achieve state-of-the-art performance on offline RL benchmarks~\citep{chen_decision_2021, janner_reinforcement_2021}. A natural question is to ask to what extent the other lessons of unsupervised pre-training can be brought to bear on sequential decision-making problems.
% \JessyComment{maybe work on pretraining-finetuning paradigm}
% \MicahComment{one big issue of the related work currently is that it hyperfocuses on RL rather than the other tasks, when we don't even mention RL in the intro and barely throughout the paper}

\prg{Transformer models.} Transformer sequence models~\citep{vaswani2017attention} have been successfully applied in other domains such as natural language processing~\citep{devlin2018bert,radford2018improving,brown2020language} and computer vision~\citep{dosovitskiy2020vit,he2021masked}. Using transformers in RL and sequential decision problems has proven difficult due to the instability of training~\citep{parisotto2020stabilizing}, but recent work has investigated how to use transformers in model-based RL~\citep{chen2021transdreamer}, motion forecasting~\citep{ngiam_scene_2021}, learning from demonstrations~\citep{recchia2021teaching}, and tele-operation~\citep{clever2021assistive}. %\citep{ngiam_scene_2021} in particular already recognized the flexibility afforded by BERT-like architectures that enable a single model architecture to be trained for multiple tasks. While they also perform multi-task training, they restrict themselves to single tasks of interest rather than training for ``all tasks'' through randomized masking. Another important difference is that their architecture is specialized for motion prediction in the autonomous driving setting, and as such only consider ``state'' predictions, rather than full sequential decision problems.

\prg{The utility of randomized masking.} In addition to being used as one of the main training objectives for BERT (the ``cloze task'', \citealt{devlin2018bert}), the flexibility afforded by randomized masking in bidirectional models has been utilized in other previous works applied to language \citep{ghazvininejad2019maskpredict,mansimov2019generalized} and vision \citep{chang2022maskgit} -- mostly for the purpose of speeding up auto-regressive decoding, which is not our focus here.

\prg{Sequential decision-making as sequence modeling.} We are not the first to consider sequential decision problems as a sequence modeling problem. \citet{chen_decision_2021} and \citet{janner_reinforcement_2021} focus on RL and show how one can use GPT-style (causally-masked) Transformer models to directly generate high-reward trajectories in an offline RL setting.
%Subsequent work has investigated how to apply this formulation to offline multi-agent problems~\cite{meng2021offline}.
Unlike this line of work, we focus on many tasks that a sequence modeling perspective enables one to do, rather than just offline-RL. While some previous work has cast doubt on the necessity of using transformers to achieve good results in offline RL~\citep{emmons_rvs_2021}, we note that offline RL~\cite{levine2020offline} is just one of the various tasks we consider.
Concurrent work to ours generalizes the left-to-right masking in the Transformer to condition on future trajectory information for tasks such as state marginal matching~\citep{furuta2021generalized} and multi-agent motion forecasting~\citep{ngiam_scene_2021}. In contrast to these works, we systematically investigate how a single bidirectional Transformer model can be trained to perform arbitrary downstream tasks (in more complex settings than motion forecasting---i.e., we also consider agent actions and rewards in addition to states). The main thing that sets us apart from these works is a systematic view of all tasks that can be represented by this sequence-modeling perspective, and a detailed investigation of how different multi-task training regimes compare.

\MicahComment{is this framing of self-supervised learning fair?}
\prg{Self-supervised learning for sequential decision problems.}
Training on random maskings of one's data can be considered as a form of self-supervised learning. Previous work on self-supervised learning is mostly focused on improving RL by  training models on auxiliary objectives such as state dynamics prediction~\citep{sekar2020planning} or intrinsic motivation~\citep{pathak2017curiosity}. % can lead to improved downstream performance.
% We note that the notion of ``task'' in the multi-task literature for RL~\citep{notorrp} typically refers to an MDP with an associated reward structure (e.g., picking up a cup vs. pouring water). However, we use ``task'' to refer more generally to an \emph{inference} or \emph{prediction task}, where RL (the task of generating trajectories that achieve high reward) is just one instantiation of the many tasks we consider (behavior cloning, backwards inference, etc.). 
Typically, to accomplish the tasks we consider, prior work relies on specialized models: for example, goal-conditioned imitation learning \citep{ding2019goal}, RL~\citep{kaelbling1993learning}, waypoint-conditioning \citep{rhinehart_precog_2019}, property-conditioning~\citep{zhan2020learning, furuta2021generalized}, or dynamics model learning~\citep{ha_world_2018, christiano_transfer_2016}.\MicahComment{https://www.nature.com/articles/s41586-019-1724-z}
In contrast, we demonstrate how sequence modeling can be a unifying framework for formulating and solving any inference task with a single model.
\MicahComment{foundation models}

\section{\fb{}}
We introduce the \fb framework, first describing how common inference tasks in sequential decision problems can be formulated as masking schemes (\Cref{sec:masking_schemes}), and then describing various possible training regimes (\Cref{sec:training}).

We model trajectories as sequences of states, actions, and return-to-go tokens:\footnote{While reward-to-go (or other trajectory statistics) are not necessary, we formulate the most general form to showcase how one can easily condition on additional properties of a trajectory if available. Using reward-to-go also enables us to baseline our method against previous offline-RL work~\citep{chen_decision_2021}.}
\[
    \tau = \{ (s_1, a_1, \hat{R}_1), \ldots, (s_T, a_T, \hat{R}_T) \},
\]
where the return-to-go $\hat{R}_t$ is the sum of rewards from timestep $t$ to the end of the episode, $\hat{R}_t = \sum_{t' = t}^T r_{t'}$.

\subsection{Tasks as Masking Schemes}\label{sec:masking_schemes}

In the \fb{} framework, we formulate tasks in sequential decision problems as input masking schemes. Formally, with the expression \textit{masking scheme} we refer to a function which randomly assigns masks to each token in trajectory snippets $\tau_{t:t+k}$ of length $k$ drawn uniformly at random from a dataset of trajectories. Each masking scheme defines both which input tokens are masked (determining what tokens are shown to the model for prediction) and which outputs of the model are masked before computing losses (determining which outputs the model should learn to predict).
%where the model must reconstruct states, actions, and/or rewards conditioned on some observations. 
% For example, conditioning on $s_{0:t}$ and $a_{0:t-1}$ and predicting $a_t$ corresponds to one inference of the behavioral cloning (BC) task: $\Prob{a_t | s_{0:t}, a_{0:t-1}}$.

\begin{figure*}[t]
  %\vskip 0.2in
  \centering
  \includegraphics[width=\textwidth]{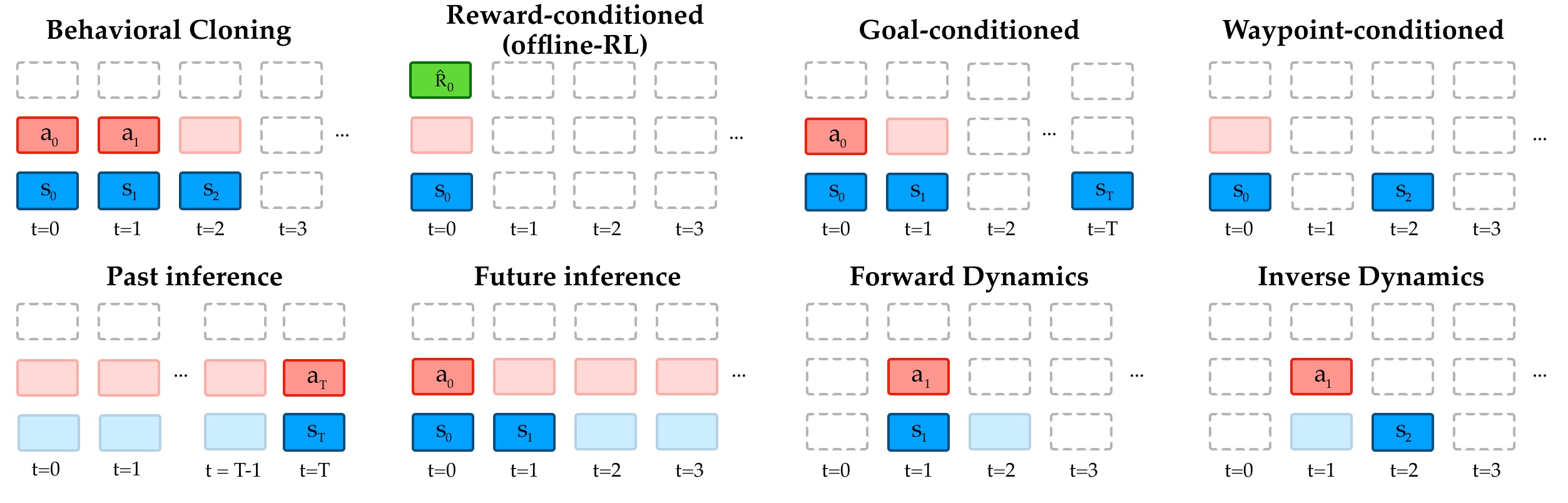}
  \vskip -0.15in
  \caption{\textbf{Just a few of the many possible tasks that can be represented in the \fb framework.} For each task we show the inputs to the model and (with lower opacity) the predictions which we train the network to make for each input masking. Note that inputs and outputs always have the same dimensionality---what differs is which tokens are masked or unmasked. For example, future inference tries to predict all future states and actions only conditional on initial states and actions. Here we only display one input masking scheme for each task, even though there might be multiple that are valid or necessary (e.g. BC will have up to T different masking schemes, one for each possible history length---although in practice one would generally use the model with a sliding window).
  }\vspace{-1em}
  \label{fig:tasks}
\end{figure*}

In \Figref{fig:tasks}, we illustrate how commonly-studied tasks such as BC, goal and waypoint conditioned imitation, offline RL (reward-conditioned imitation), and dynamics modeling can be unified under the representation of tasks as masking schemes. We describe the masking scheme for each of these tasks below.

For each trajectory snippet $\tau_{t:t+k}$ in each batch:

\begin{itemize}
    \item \textbf{Behavioral Cloning.} Select $i \in [0, k]$ uniformly. Feed $s_{t:t+i}, a_{t:t+i-1}$ to the network (include no actions if $i=0$), with all other tokens masked out. Have the network only predict the next missing action $a_i$.
    % \item \textbf{Next action completion.} Uniformly at random, sample from the dataset an action to predict $a_t$. Give the network the context $s_{t-k}, \dots, s_t, a_{t-k}, \dots, a_{t}$, with $a_t$ masked out. If the action is at the beginning of a trajectory, mask the all tokens in the extra context length. This is equivalent to standard behavior cloning training. 
    \item \textbf{Goal-Conditioned imitation.} Same as BC, but $s_{t+k}$ is always unmasked.
    \item \textbf{Reward-Conditioned imitation (Offline-RL).} Same as BC, but return-to-go $\hat{R}_t$ is always unmasked.
    \item \textbf{Waypoint-Conditioned imitation.} Same as BC, but a subset of intermediate states are always unmasked as waypoints or subgoals. %$s_{t_1}, \ldots s_{t_n}$ are always unmasked.
    \item \textbf{Future inference.} 
    % Uniformly at random, sample a sequence of length $k$ from all data. 
    % Select $t \in [0, k]$ uniformly. Mask the end of the trajectory $\{a_t, s_{t+1}, a_{t+1}, \ldots, a_w, s_w\}$ where $w$ is the last index of the context window. 
    Same as BC, but the model is trained to predict all future states and actions, rather than only the next missing action.
    % For each sequence, we generate $T$ training sequences $\tau_{i=1}^T$. For training sequence $\tau_i$, we mask the end of the trajectory $\{(a_t, s_{t+1}, a_{t+1}, \ldots, a_T, s_T\}$.
    \item \textbf{Past inference.} Select $i \in [1, k]$ uniformly. Feed $s_{t+i:t+k}, a_{t+i:t+k}$ to the network, with all other tokens masked out. Have the network predict all previous states and actions $s_{t:t+i-1}, a_{t:t+i-1}$.
    %We condition the model on a final state $s_{T-1}$. 
    %We roll out predicted actions \emph{and} states in reverse: for each timestep $t \in (T-2),\ldots, 0$, first sampling $a_t = \arg\max_{a_{t}'} p(a_{t}' \mid s_T, a_{T-1}, s_{T-1}, \ldots, s_{t+1})$ . Since we do not typically have access to an inverse dynamics model, we query the model for the state at time $t$ given the future trajectory: $s_t = \arg\max_{s_{t}'} p(s_{t}' \mid s_T, a_{T-1}, s_{T-1}, \ldots, s_{t+1}, a_{t})$. \MicahComment{make probability notation consistent across paper}\MicahComment{give more details about waypoinging in gridworld later}
    \item \textbf{Forward dynamics.} Select $i \in [0, k-1]$ uniformly. Give the network the current state and action $s_{t+i}, a_{t+i}$, and have it predict the next state $s_{t+i+1}$. In theory, this could enable to handle also non-Markovian dynamics (we did not test this).
    \item \textbf{Inverse dynamics.} Select $i \in [1, k]$ uniformly. Give the network the current state and previous action $s_{t+i}, a_{t+i-1}$, and have it predict the previous state $s_{t+i-1}$.
    \item \textbf{All the above (ALL).} Randomly select one of the above masking scheme functions and apply it to the current sequence. This is a simple way of performing multi-task training.
    \item \textbf{Random masking (RND).} 
    % Uniformly at random, sample a sequence of length $k$ from the data. 
    Randomly select the masking probability for the sequence $p_{\text{mask}} \sim \text{Uniform}(0,1)$. Mask each token independently with probability $p_{\text{mask}}$  (more information about this masking scheme is in \Cref{sec:rnd-masking}). Have the model predict all tokens that were masked in the input (excepts returns-to-go). We treat the return-to-go $\hat{R}_0$ as a special token, masking it from the input with probability $p_{\text{mask}} = 0.5$. Randomly using the return-to-go in this fashion enables the model to perform both reward-conditioned and non-reward-conditioned tasks at inference time.
\end{itemize}

\subsection{Model Architecture \& Training}\label{sec:training}
\begin{figure}[t]
\vskip -0.2in
\begin{center}
\includegraphics[width=0.4\textwidth]{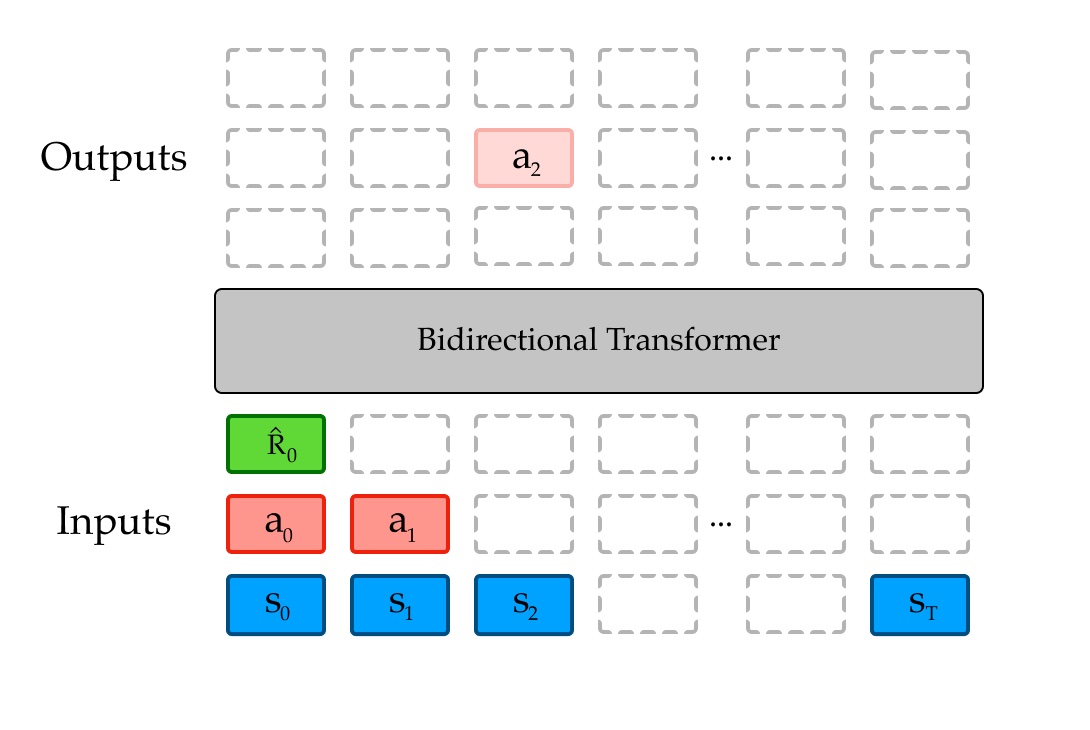}
\vskip -0.15in
\caption{The \fb{} model takes in a snippet of a trajectory which is masked according to a masking scheme before inference time. For each input possible masking, there are (many) corresponding tasks of predicting the missing inputs. Above we show an input masking corresponding to conditioning on both reward \textit{and} final (goal) state; we highlight the output corresponding to predicting the agent's next action, i.e. performing the inference $\Prob{a_2 | s_{0:2, T}, a_{0:1}, \hat{R}_0}$.}
\label{fig:fb-model}
\end{center}
\vskip -0.1in
\end{figure}

As model architecture, we use stacked bidirectional transformer encoder (self-attention) layers, similarly to BERT \citep{devlin2018bert}. See \Figref{fig:fb-model} for a visual representation of usage, and \Cref{appx:model_details} for more details.

Given all of the possible masking schemes with which one could train \fb{} models, there will be many reasonable approaches to training and using such model at test. Below, we consider some, weighing their pros and cons:

\begin{enumerate}
    \setcounter{enumi}{-1}
    \item \textbf{Training specialized models (e.g. BC)}: using a single \fb{} model for a single task. This involves training on only one of the single-task masking schemes from \Cref{sec:masking_schemes} (except ALL and RND, which are multi-task by nature). The only advantage that this approach might have over conventional model types is the ease of use of tapping into the same architecture for every task instead of designing task-specific input pipelines; this is not our main focus.
    \item \textbf{Training on multiple pre-specified useful tasks (e.g. ALL)}: with this setup, we aim to train a single \fb{} model that can perform well on all of its training tasks. % the goal in this case would be to obtain a single \fb{} model which can do a decent job at all the tasks it was trained on;
    % and perhaps some new (similar) ones not seen at train time
    We expect this setup to often perform similarly or better than custom models, as \fb{} may learn structure in the data from the other masking schemes that enables it to surpass the performance of the customized models.
    %we'd expect this to perform on par with specialized models for any training task, and maybe sometimes slightly better because (much like in language) we can hope to learn structure in the data from the other masking schemes used in training;
    \item \textbf{Training on random \maskings (RND)}: with this approach, we aim to train a single model that performs well on any arbitrary sequence inference, without having to specify the tasks of interest before training. 
    %the goal of this approach is for it to lead to a single model performing well on any arbitrary sequence inference (not requiring to pre-specify tasks of interest before training). 
    Moreover, this could potentially improve representations compared to (1), as the network is forced to reason about all components of the environment in order to perform arbitrary prediction tasks. A disadvantage compared to (1) or even (0) is that depending on the model context length, the space of possible \maskings could become too large for the model to be able to learn all tasks well;
    \item \textbf{Training with methods (1) or (2), and then fine-tuning to a test-time task}: this should enable to take advantage of the improved representations from multi-task training, and specialize the model parameters to the single task at hand (without having to share model capacity with other tasks). Therefore, this approach should lead to the best of both worlds (in terms of performance) of multi and single-task models. The main cost would be that of greatly diminishing the multi-task test-time flexibility enabled by (1) or (2).
\end{enumerate}
% \adnote{cool, we have this framework, what do we do it? (consier calling flexibit a framework, and different models emerge out of different ways of training it); we can train models in various ways, each with potential advantages of convential architectures:

% 2. train on random maskings (only, or in addition to prespecified tasks); % 3. do 1 or 2, but fine-tune on the test task; advantage:..;

\prg{Hypotheses.} Based on the above observations, we hypothesize that:
\textbf{H1.} (3), which uses the full power of the framework, is often better than (0); \textbf{H2.} (1) and (2) are sometimes better than (0); \textbf{H3.} when masking coverage of the desired tasks is sufficient, (2) is better than (1); \textbf{H4.} (3) is slightly better than (1) and (2);

In \Secref{sec:gridworld}, we first show how training with a random masking scheme enables a single \fb{} model to perform arbitrary inference tasks at test-time on a simple navigation environment dataset, without the need for specialized output heads or training schemes that are customized for the downstream task. We then additionally show how on specific tasks of interest, training with random \maskings can achieve comparable or better performance to specialized models or multi-task models (trained on a short list of possibly relevant tasks), despite rarely seeing that particular task at training time. %, the masked prediction task enables \fb{} to outperform specialized models by serving as a form of data augmentation. 
After training on the masked prediction task, we show how \fb{} can additionally be fine-tuned on the specific task of interest, which we show further improves performance.

% In \Secref{sec:complex-envs}, we present preliminary experimental results from more complex robotics environments, showcasing that performing random-masking training and fine-tuning can obtain comparable or better reward performance in the behavior cloning task.
\section{Experiments}
\label{sec:gridworld}

\subsection{Experimental Setup}

To showcase the flexibility of our model, we set up a grid-like navigation environment in which we perform many different types of inferences. We use the minigrid environment framework \citep{chevalier2018minigrid} with a custom DoorKey setting. Our environment is a fully observable 4-by-4 gridworld in which the agent should move to a fixed goal location which is behind a locked door. To be able to pass the locked door, the agent must first pick up a key. Both the agent and key are initialized at random locations in each episode outside the locked room with the goal. The agent receives a reward of $1$ for each timestep it moves closer to the goal, $-1$ if it moves away from the goal, and $0$ otherwise. We train \fb{} models on training trajectories of sequence length $T=10$ from a noisy-rational agent \citep{ziebart_maximum_nodate} which takes the optimal action most of the time, but has some chance of making mistakes proportional to their sub-optimality. More specifically, the agent takes the optimal action with probability $a \sim p(a) \propto \exp (C(a))$ where $C(a)=1$ if the distance to the current goal (key or final goal) decreases, $-1$ if it increases, and $0$ otherwise. %We train on the random masking task described in \secref{sec:pretraining} \todo{check that all relevant details are described there}.
More detailed information about the environment is in \Cref{sec:doorkey-details}.

\subsection{One model to rule them all}\MicahComment{inconsistencies in capitalization for subheadings}

\begin{figure*}
    \vskip -1em
    \centering
    \includegraphics[width=\textwidth]{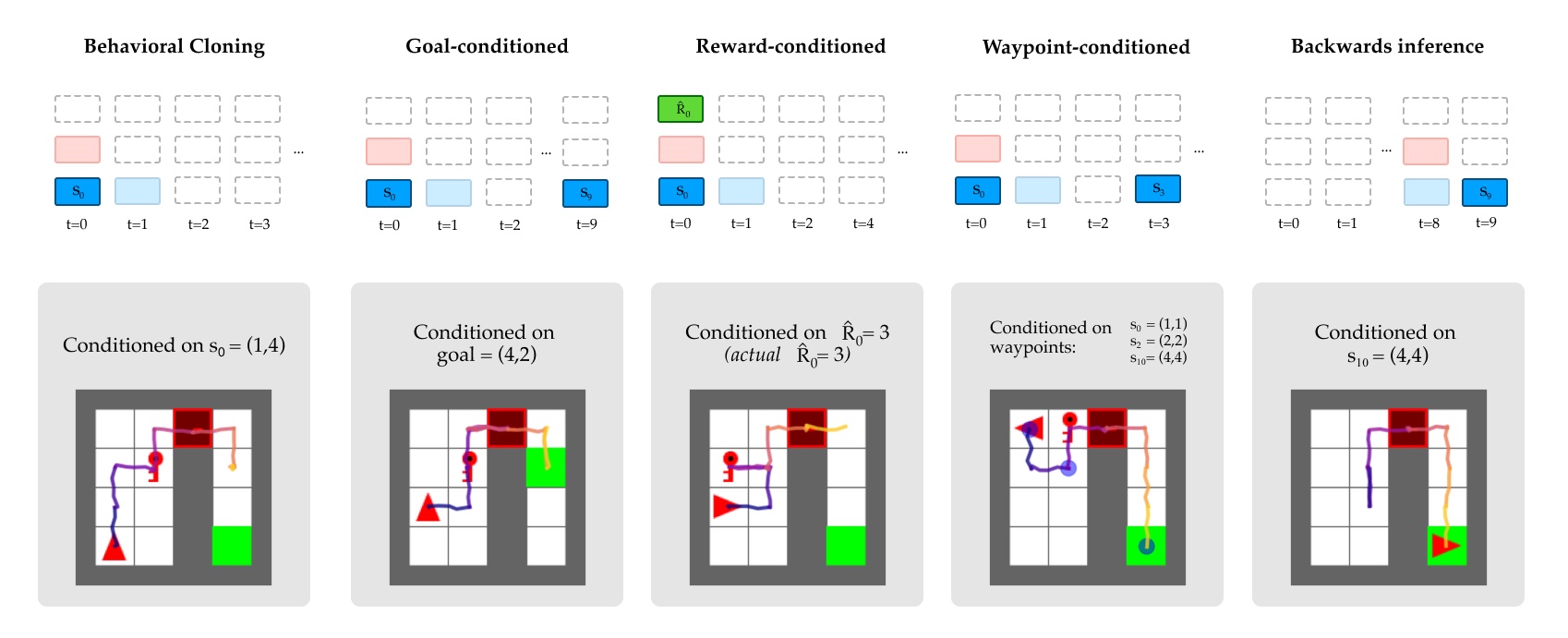}
    \vskip -1.5em
    \caption{\textbf{A \fb model trained with random masking queried on various inference tasks.} \textbf{(1) Behavioral cloning:} generating an expert-like trajectory given an initial state. \textbf{(2) Goal-conditioned:} reaching an alternative goal.
    %, then the original goal (4,4) and finally navigate to state (4,2) for the last timestep. 
    \textbf{(3) Reward-conditioned:} generating a trajectory that achieve a particular reward, e.g., by taking a suboptimal path that does not directly reach the goal. \textbf{(4) Waypoint-conditioned:} reaching specified waypoints (or subgoals) at particular timesteps, e.g. going down on the first timestep instead of immediately picking up the key. \textbf{(5) Backwards inference:} generating a likely \emph{history} conditioned on a final state (by sampling actions and states backwards). Trajectories are shown with jitter for visual clarity. \MicahComment{coloring of traj}}
    \label{fig:gridworld-results}
    \vskip -1em
\end{figure*}

We first showcase the out-of-the-box flexibility of a \fb{} model when trained with random masking. By querying the model in a variety of ways, we show qualitatively in \Figref{fig:gridworld-results} that it is able to perform common tasks. Unless otherwise indicated, we take the highest probability action from the model $a_t = \arg\max_{a_t'} p(a_t' \mid s_0, a_0, \ldots, s_t)$, and then query the environment dynamics for the next state $s_{t+1}$. In some tasks, trajectories rolled out from the model may not always be consistent with the observed factors (e.g. the model's highest probability trajectory may not reach the conditioned waypoints). In these cases, one could use a search procedure such as beam search to expand several trajectory hypotheses, eliminating the ones that are inconsistent with observed factors, although we did not find this to be necessary in the gridworld examples. We use a simpler procedure explained in \cref{sec:doorkey-details}.

As seen in the backwards inference task (where we also query the model for state predictions), we can also effectively use \fb{} as an inverse dynamics model; in our framework, this is simply yet another task captured by the corresponding masking scheme.

\todo{weave in some discussion of how often it is successful?}
\todo{check indices are all 0-indexed everywhere}

Note that with random masking, seeing the exact masking corresponding to a particular task at training time is exceedingly rare; there are  $2^T * 2^T * 2$ possible state, action, and reward maskings for a sequence of length $T$ (in these experiments $T=10$, resulting in about $2*10^6$ possible maskings). This suggests that the model is capable of generalizing across input-masking schemes.

\subsection{Future State Predictions}
\begin{figure*}[t]
\noindent
\begin{minipage}[!t]{0.35\textwidth}
    \centering
    \includegraphics[width=\textwidth]{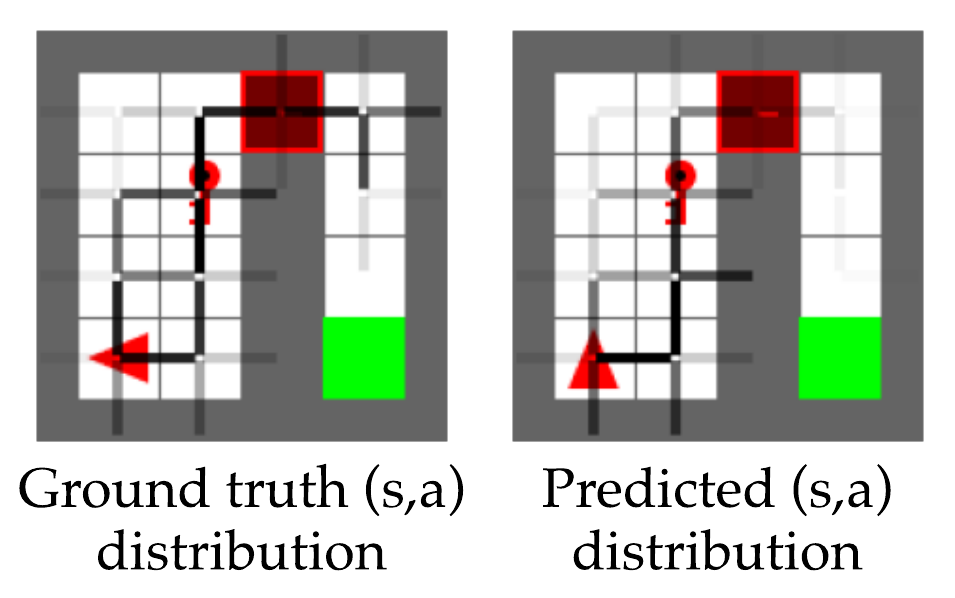}
    \vskip -1.3em
    \caption{Distribution of states and actions for trajectories in the validation set, vs. trajectories sampled from the model, conditioned on the initial agent position (1,4) and key position (2,2).}
    \label{fig:sa-viz}
\end{minipage}%
\hfill%
\begin{minipage}[!t]{0.55\textwidth}
    \centering
    \includegraphics[width=\textwidth]{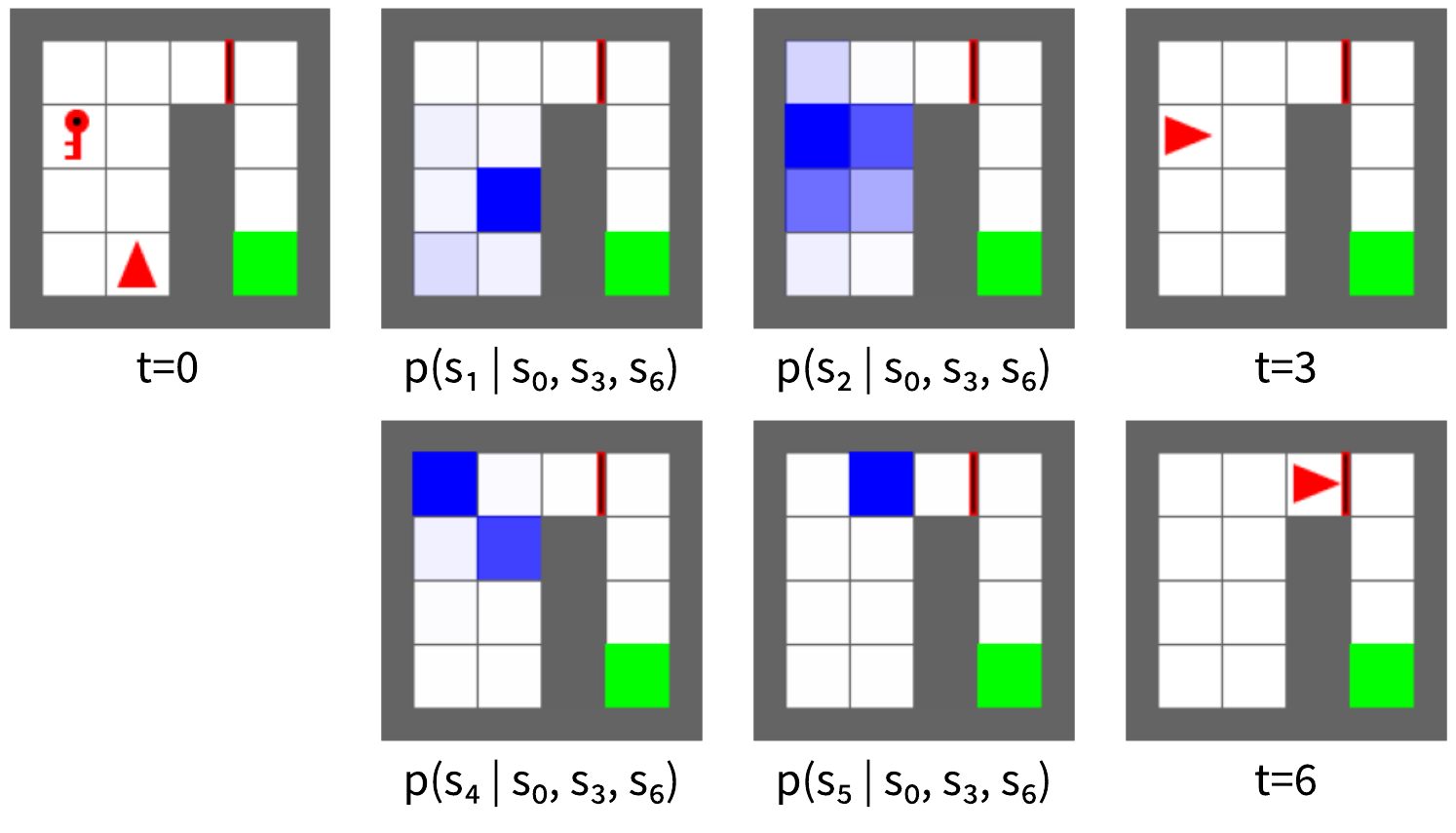}
    \vskip -1.3em
    \caption{Predicted state distributions, conditioned on states at $t=0, 3, 6$.}
    \label{fig:state-viz}
\end{minipage}%
\end{figure*}
In \Figref{fig:gridworld-results} we show how we can iteratively use the randomized-masking \fb{} model to sample trajectories in the environment. We visualize the distribution of states and actions for trajectories sampled from the model, conditioned on the initial state (essentially, looking at the transition frequencies of BC-sampled trajectories). As seen in \Figref{fig:sa-viz}, the model learns to match the underlying distribution of trajectories of the agent (as can be verified by comparing to held-out data). 
 
Uses of the random-masking-trained \fb{} model are not limited to the \textit{iterative sampling} of new trajectories (requesting inferences about the agent's next action). One can also request inferences for states and actions further into the future: e.g., ``where will the agent be \textit{in 3 timesteps}?''. Given a \emph{fixed} initial set of observed states, we visualize the distribution of predicted states at each timestep in \Figref{fig:state-viz}. Since we do not roll out actions, querying the model for the predicted state distribution at a particular timestep marginalizes over missing actions; for example, $\Prob{s_1 \mid s_0, s_3, s_6}$ models the possibility that the agent chooses either up or left as the first action. Looking at these predictions, we see that most of the state predictions made by the model seem plausible, suggesting that it is able to appropriately leverage its knowledge of dynamics and usual agent behavior. In particular, it correctly models that the agent has equal probability of going up and right at $t=3$ (leading it to the distribution over states at $t=4$), and that the agent must be at position (2, 1) at $t=5$ in order to reach the door at $t=6$.

\subsection{How do different training regimes compare?}

\begin{figure}
    \centering
    \includegraphics[width=0.7\columnwidth]{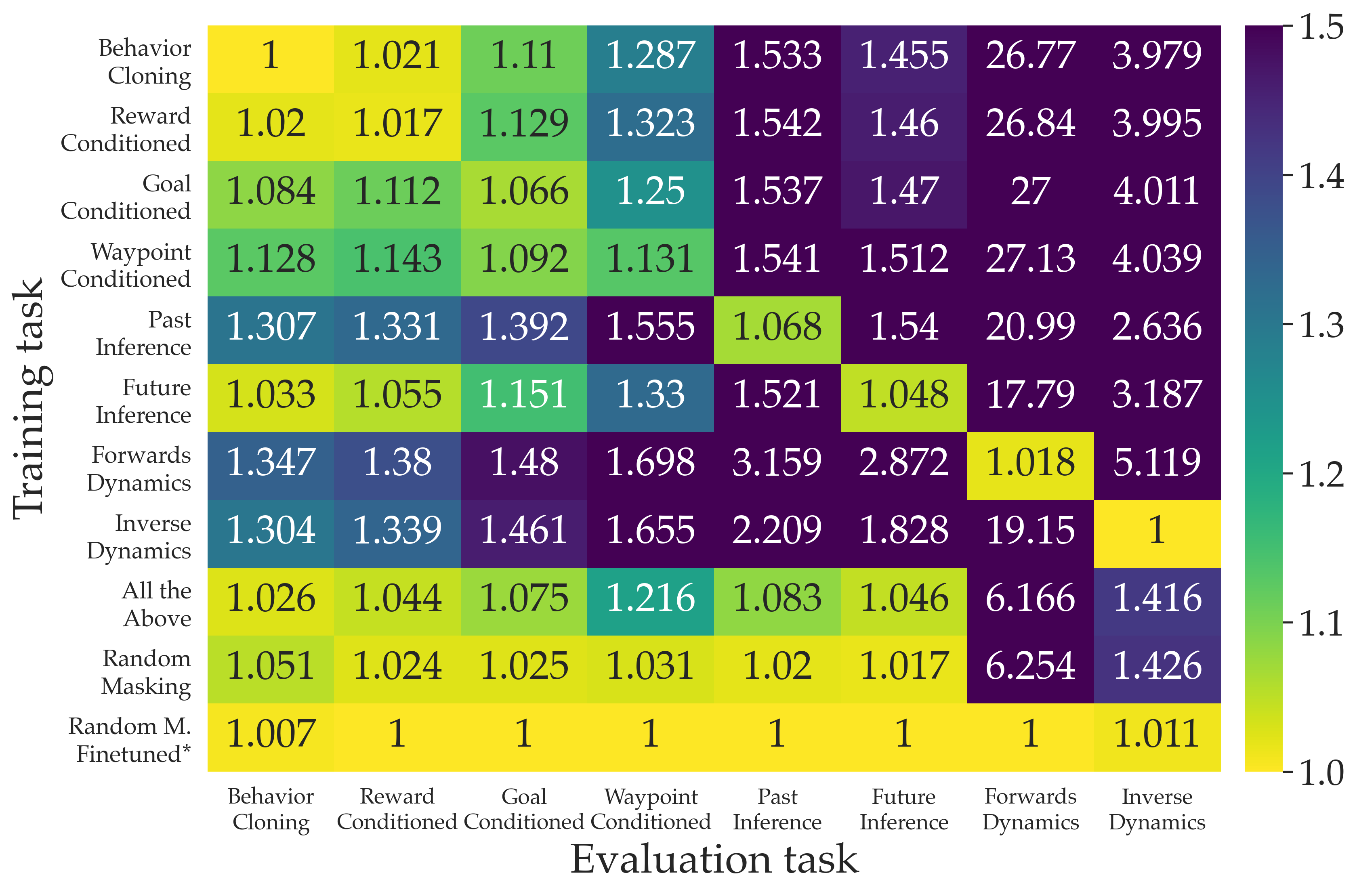}
    \caption{\textbf{Task-specific validation losses (normalized column-wise).} Each row corresponds to the performance of a single model evaluated in various ways, with the exception of the last row---for which each cell is fine-tuned on the respective evaluation task. Loss values are averaged across 3 seeds and then divided by the smallest value in each column. Thus, for each evaluation task (i.e., column), the best method has value $1$; a value of 1.5 corresponds to a loss that is 50\% higher than the best model in the column. See \Cref{sec:more-validation-minigrid} for unnormalized values.}% \sm{can we use wrapfig for this figure? or is it banned}\MicahComment{i believe we can use it -- go ahead and do it if you want!}}
    \MicahComment{have a 'measuring performance prg'}
    \label{fig:gridworld-valid-ranks500}
\end{figure}

If we care about a single task (e.g. goal-conditioned imitation), should we train a model simply on that task? Or can there be advantages to training a general model first, and then fine-tuning it to the task of interest? To shed light on these questions we compiled a heatmap (\Figref{fig:gridworld-valid-ranks500}) which reports validation losses across a variety of training and evaluation tasks.
Our experimental findings are summarized as follows:

\prg{Single-task models do well on their task and poorly on others.} Firstly, as a simple sanity check, notice that all entries on the diagonal (the specialized models, trained and evaluated on a single task) are among the best at the task (that is, their rescaled loss values are quite low relative to other entries in their column). Given that they also use the \fb model architecture, they can also be evaluated on other tasks that they weren't trained on (just by changing the input maskings): generally, as one would expect, these single tasks model tend to suffer quite a bit of degradation in performance when evaluating on previously unseen evaluation tasks. %One exception is behavior cloning and offline-RL (behavior cloning), which seem to only lose a couple percentage points: this is likely an indication that in this environment, the reward information is not very informative

\prg{Multi-task models perform as well as specialized ones (and sometimes better).} Turning our attention to the third and second-to-last rows, we see multi-task models: for both models, we have a single model that is able to reasonably perform across all the evaluation tasks considered. In the case of the randomized masking, there are two notable takeaways: 1) it leads to lower loss values across almost all evaluation tasks relative to the ALL model, giving credence to the idea that it might be beneficial to broaden the number of tasks trained on dramatically---even if one is interested in a single task (this supports \textbf{H3}); 2) often the random-masking \fb{} model is able to (even before fine-tuning) obtain better performance than the specialized models (in 4/8 of the tasks)---supporting \textbf{H2}; 3) we see that this proportion increases to 6/8 tasks after additionally fine-tuning this general \fb{} model to the specific task of interest (supporting \textbf{H1} and \textbf{H4}).

\prg{Forward and inverse dynamics.} The forward and inverse dynamics evaluations of multi-task models stick out as being particularly poor performance relative to specialized and fine-tuned models. This is because this simple environment has deterministic dynamics, meaning that with sufficient data (as in this case), it's essentially impossible to overfit (``overfitting is fitting''). Training a model for long enough exclusively on forward dynamics gives loss values quite close to zero. This means that any multi-task model which is trading off between that task and others might fail to overfit quite as dramatically (a gap which can be easily made up for during fine-tuning). In this sense, it might be more appropriate to consider the heatmap without these tasks. Still, we wanted to demonstrate how well our model could perform these tasks, if necessary.

\MicahComment{eventually add the below paragraph back in}
% \prg{Overfitting in specialized models.} One thing that we noticed by looking at the training curves\MicahComment{eventually include} is that specialized models tended to overfit much quicker than the random-masking-trained model -- which sees the same sequence of data every time with drastically different maskings (potentially constituting an effective form of data-augmentation). This suggests that the pre-fine-tuning benefit of multi-task training might be reduced or increased by respectively increasing or reducing the amount of data. We used 10x less data, and 2x more data, but didn't see much difference in the model orderings, warranting further investigation. These additional results are in \appref{}\MicahComment{todo} 

\MicahComment{eventually we could actuad a demo experiment where we try to infer the distribution of the location of the key at t=0 with a past-prediction model}
% \subsection{How does this single model compare to specialized ones?}

% \prg{High-data regime: similar or better performance after fine-tuning.} If we fine-tune we can do as well or slightly better than specialized models which were trained on the individual task. Any improvement is simply due to having better representations, gained from training on the extra tasks.

% \prg{Low-data regime: better performance out-of-the-box.} This is thanks to the data augmentation effect. We see that the gap becomes larger.

% \MicahComment{We could add a experiment for this?} We prevent a specific type of masking from ever appearing at training time, and the model still does well.

% \subsection{When can we expect better prediction than alternative models?}

% The main avenues through which one would expect \fb{} to perform better than alternatives is through 1) better representations obtained by multi-task training, and 2) less overfitting, as enabled by the data augmentation effect of multi-task training.

% In our experiments, it seems like 2) is playing a bigger role than 1) [provide evidence]

% \input{sections/50-complex-tasks}

\MicahComment{fix author list (steph affiliation)}

\section{Limitations and Future Work}

% \prg{Offline RL.} Although offline-RL is used in this paper just as a means to display our model's flexibility, confronting the issues in the Mujoco environments described in \secref{} made us confront various questions: what do we intend our learners to do when they are presented with a very small amount of data (e.g. a single trajectory) to learn from? If environments are amenable to achieving high reward through memorization/regurgitation of expert trajectories - should we steer away from this path or embrace it? Do we have expectations that an offline RL method will be able to generalize beyond the set of states it has seen in the training data? \MicahComment{it's not just about this -- also in general: are low randomness environments good indicators of performance? are any of the comparisons we gain in that domain meaningful, or just capture a model's capability of memorizing/regurgitating data?}

% Hyperparameter Selection for Imitation Learning - imitation learning MSE / reward.

\prg{More complex environments.} An important limitation of our work is that our results are currently limited to a simple minigrid domain. A clear axis of improvement for our results would be to scale the experiments to more complex environments, which would enable better testing of the limits of our methods.

\prg{Property-conditioning.} The choice of representing trajectories as states, actions, and returns-to-go is not constrained by the model. One could use any property of the environment or the agent as additional tokens (or set of tokens), enabling to perform property-conditioned inferences, or infer such properties from test-time trajectory snippets. Reward-conditioning is just one example of such property-conditioning \citep{zhan2020learning, furuta2021generalized}.

\prg{Short context lengths.} One limitation in our experimentation are the relatively short context lengths used. This is because 1) the architecture we use requires memory quadratic in the sequence length, and 2) the number of possible random maskings is exponential in the sequence length.
% In addition to reducing the expressivity of the model, using short sequence lengths reduces the applicability of maskings such as goal and waypoint conditioning (which can only condition on future states within the context window).
1) could be addressed by using efficient transformers~\citep{wang2020linformer, jaegle2021perceiver}; 2) could be addressed by designing masking schemes tailed to specific test-time tasks (see \Cref{sec:rnd-masking}), or using a more principled masking schemes \citep{levine2020pmi}.

\prg{Other future work.} Another exciting direction for future work is trying to see whether the benefits obtained from random masking (or even multi-task masking) would also apply to Bayes Networks more generally; alternatively, even trivially extending the approach to multi-agent settings (for which token-stacking could prove more valuable), could enable many interesting masking-enabled queries \cite{ngiam_scene_2021}. %More work on understanding what information BERT learns, and how it is stored, and whether and how multi-task training improves representations in the setting of sequential decision problems also seems warranted \citep{Rogers}.

% Additionally, we propose a flexible way of handling inputs that enables us to substantially increase the sequence length that is feasible to handle by our model relative to previous work \citep{chen_decision_2021, janner_reinforcement_2021}.

% Syntetic demonstrations are not necessarily good proxy for human data (sec 5 \citep{orsini2021matters})

% 

% \section{Conclusion}

\section{Conclusion}
In this work we propose \fb, a framework for flexibly defining and training models which: \textbf{1)} are naturally able to represent any inference task and support multi-task training in sequential decision problems, \textbf{2)} match or surpass the performance of specialized models after multi-task pre-training, and almost always surpasses them after fine-tuning. We believe our approach provides an elegant conceptual formulation and exciting experimental results which warrant further investigation by the research community.

\MicahComment{replace all mentions of IL/imitation learning with BC/behavior cloning}

\newpage

\subsubsection*{Acknowledgments}
We'd like to thank Miltos Allamanis, Panagiotis Tigas, Kevin Lu, Scott Emmons, Cassidy Laidlaw, and the members of the Deep Reinforcement Learning for Games team (MSR Cambridge), the Center for Human-Compatible AI, and the InterAct Lab for helpful discussions at various stages of the project. This work was partially supported by Open Philanthropy and NSF CAREER.

\bibliography{refs}
\bibliographystyle{gpl_iclr2022_conference}

\newpage
\appendix
\section{Appendix}

\subsection{Model details}\label{appx:model_details}

% \todo{discuss embedding sizes and other model details that are diff from transformer, stacking, how we only ever condition on return-to-go at timestep 0}

\prg{Input stacking.} %The main bottleneck for transformer models is usually the input sequence length: computational cost grows $O(n^2)$. In the context of decision processes, stacking state, action, and reward inputs would trivially enable to reduce the number of sequential inputs to the transformer by $3$ times relative to DT \citep{chen_decision_2021} and by $(N+M+1)$ times relative to Trajectory Transformer (where $N,M$ are the dimensionality of the state and action spaces) \citep{janner_reinforcement_2021}.
An important hyperparameter for transformer models is what dimension to use self-attention over. Previous work applies it across states, actions, and rewards as separate tokens (or even individual state and action dimensions) \citep{chen_decision_2021, janner_reinforcement_2021}; this can increase the effective sequence length that we would need to input a trajectory snippet of length $k$: for example if treating states, actions, and rewards separately (have self-attention act on each independently), the sequence length would be $3k$. While this is not an issue for the short context windows we use in our experiments, this seems wasteful: the main bottleneck for transformer models is usually the computational cost of self-attention, which scales quadratically in the sequence length. To obviate this problem, we stack states, actions, and rewards for each timestep, treating them as single inputs. This way, we are making self-attention happen only across timesteps, reducing the self-attention sequence length required to $k$. This also seems like a potentially advantageous inductive bias for improving performance. Though we did not test this systematically, preliminary experiments did show that input stacking reduced validation loss.

\prg{Return-to-go conditioning.} Unlike previous work that considers return-to-go conditioning \citep{chen_decision_2021, janner_reinforcement_2021}, we don't provide the model with many return-to-go tokens (one for each timestep). Providing just the first token should be sufficient for the model to interpret the return-to-go request (as, if necessary, the model can compute the remaining return to go in later steps). We found that this did not negatively impact performance, either for single-task reward-conditioned training, or for randomized masking training.

% \prg{Positional/timestep encoding.} \MicahComment{We don't use timestep encoding anywhere in the paper as of right now, as it's not necessary for the minigrid bc of the horizon being short and also for mujoco (when not using RC)}

\subsection{Random masking scheme details}\label{sec:rnd-masking}
Given the correspondence of maskings and tasks, we found that a masking scheme that uniformly masks each token with a fixed probability $p$ (as common in NLP) to be a naive strategy, particularly for longer sequence lengths. With this approach, the number of tokens masked will be distributed as $n \sim Binomial(N, p)$ (where $N$ is the number of tokens). This means that it will be highly unlikely to see either $0$ or $N$ tokens masked -- even though there are many meaningful tasks that require most tokens to be masked (past prediction) or unmaksed (behavior cloning with full history). We fix this problem by using the randomized masking scheme described in \Cref{sec:masking_schemes}, which we found to perform much better.

\subsection{Minigrid details} \label{sec:doorkey-details}

Below we delineate some more details about our custom DoorKey minigrid environment.

\prg{State and action space, and more details on environment dynamics}

The state and action spaces are both represented as discrete inputs: there are 4 actions, corresponding to the 4 possible movement directions (up, right, down, left); taking each action will move the agent in the corresponding direction unless 1) the agent is facing a wall, 2) the agent is facing the locked door without a key. Stepping on the key location tile picks up the key. The state is represented as two one-hot encoded position vectors---the agent position and the key position (which is equivalent to the agent position once the key has been picked up). Both agent and key position have 16 possible values, some of which are never seen in the data (e.g. the agent position coinciding with a wall location). Together, such vectors are sufficient to have full observability for the task---as seeing the key location coincide with the agent location informs the model that the agent is holding the key, and if the agent is holding the key it can open the locked door. Once the agent is holding the key, whether the door is open or closed is irrelevant.

Having the states and actions be discrete enables all model predictions to be done on a discrete space---which is particularly convenient as it enables the trained models to output any distribution over predicted states and actions, which can be easily visualized such as in \Cref{fig:state-viz}.

\prg{\fb models hyperparameters for DoorKey experiments}

As the DoorKey environment have discrete actions and states, we use the softmax-cross-entropy loss over all predictions.

For the trained models we use 3 stacked encoder layers, 8 attention heads, 128 hidden dimension size, and 128 embedding dimension size. We train with a learning rate of $10^{-4}$, and but scale the state and action losses by a factor of $0.2$. We train with the torch implementation of the Adam optimizer. We use batch size 100. We train until we reach the lowest validation loss, or up to a maximum of 6000 epochs. 

When fine-tuning, we reduce the learning rate to $5 \times 10^{-5}$ and train for 500--1500 epochs depending on the task.

\prg{Fixing prediction inconsistencies}

In backwards inference, we note that sometimes the predicted state at the previous timestep may not be consistent with the dynamics of the environment or the observed states. In cases where the prediction is inconsistent with environment dynamics, we re-sample the prediction (rejection sampling). In cases where the prediction is inconsistent with the observed variables, we simply return the trajectory even though it may not be consistent with the conditioned states, although rejection sampling can also be performed here.

\subsection{More validation-loss results in the minigrid domain.}\label{sec:more-validation-minigrid}

We report below more validation-loss results from the minigrid experiments, varying the amount of data used to train all models. We try dataset sizes of 50, 500, and 1000. We see that notwithstanding the differences in dataset size, the trends and relative orderings of performance between models tend to stay the same.

\begin{figure}[h!]
    \centering
    \includegraphics[width=0.7\columnwidth]{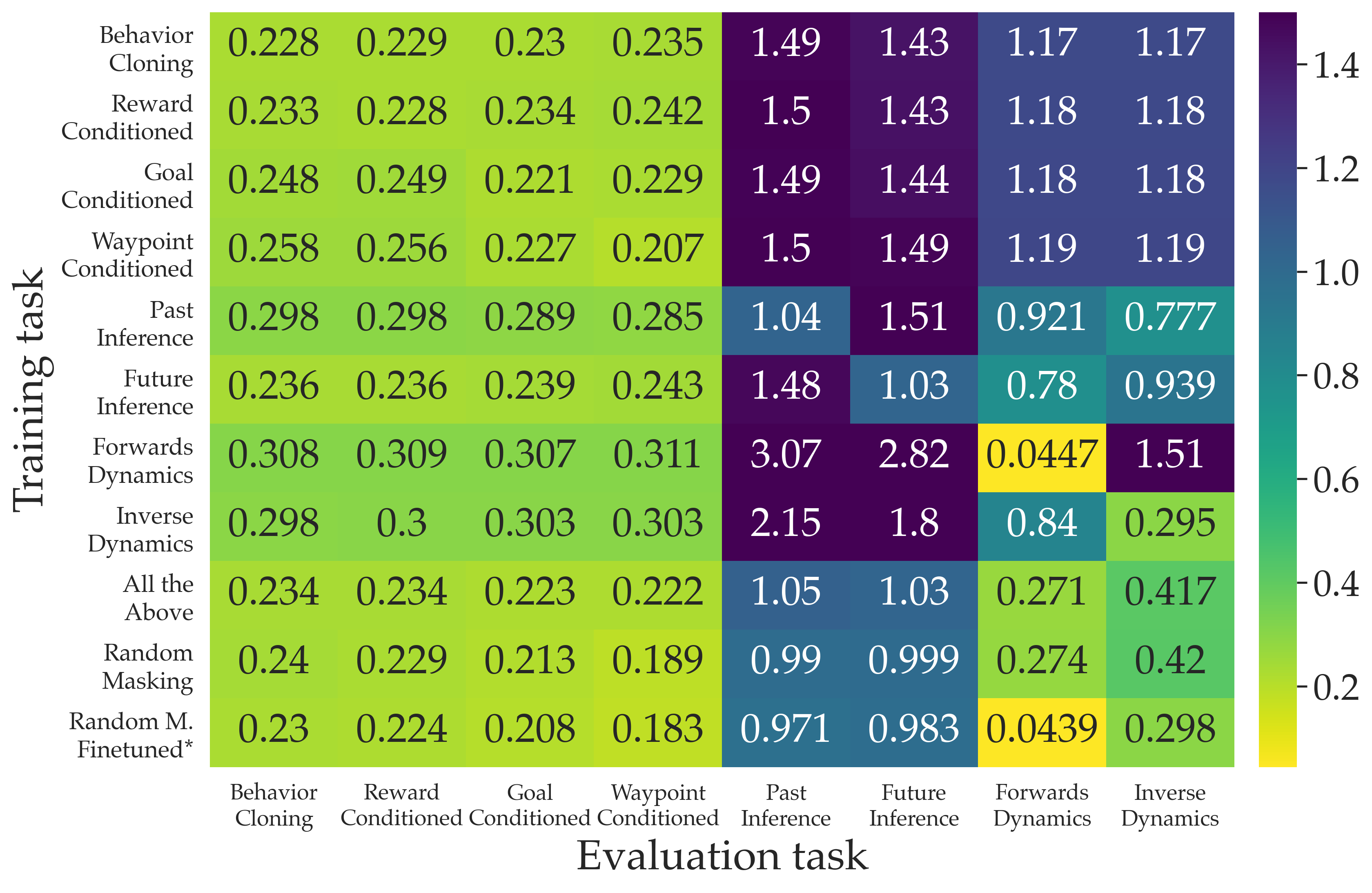}
    \caption{The raw loss values corresponding to \Figref{fig:gridworld-valid-ranks500}.}
    \label{fig:gridworld-valid-raw500}
\end{figure}
\begin{figure}[h!]
    \centering
    \includegraphics[width=0.7\columnwidth]{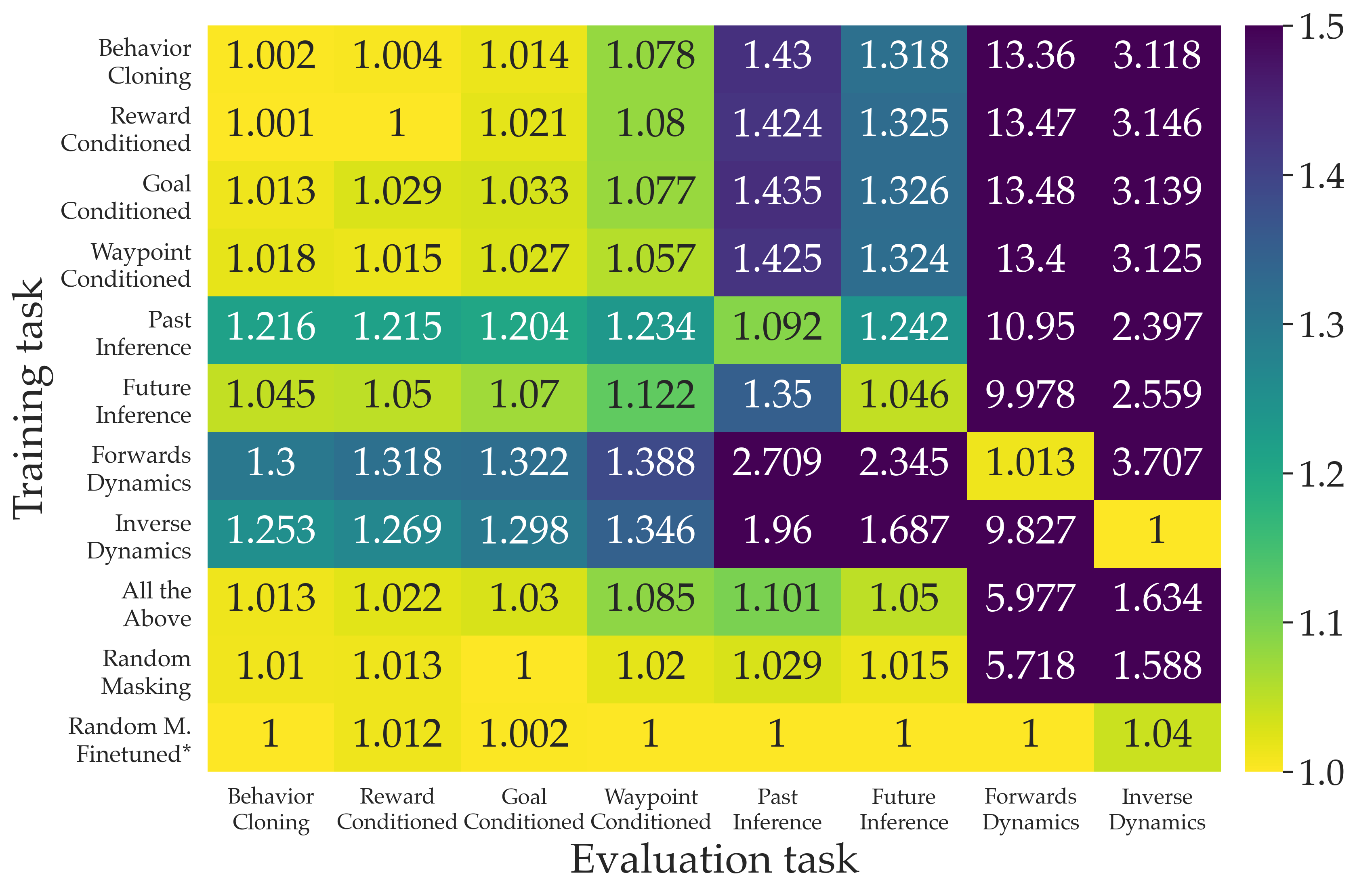}
    \caption{\Figref{fig:gridworld-valid-ranks500} when using a dataset of 50 trajectories instead of 500. We see that most of the trends remain the same.}
    \label{fig:gridworld-valid-ranks50}
\end{figure}
\begin{figure}[h!]
    \centering
    \includegraphics[width=0.7\columnwidth]{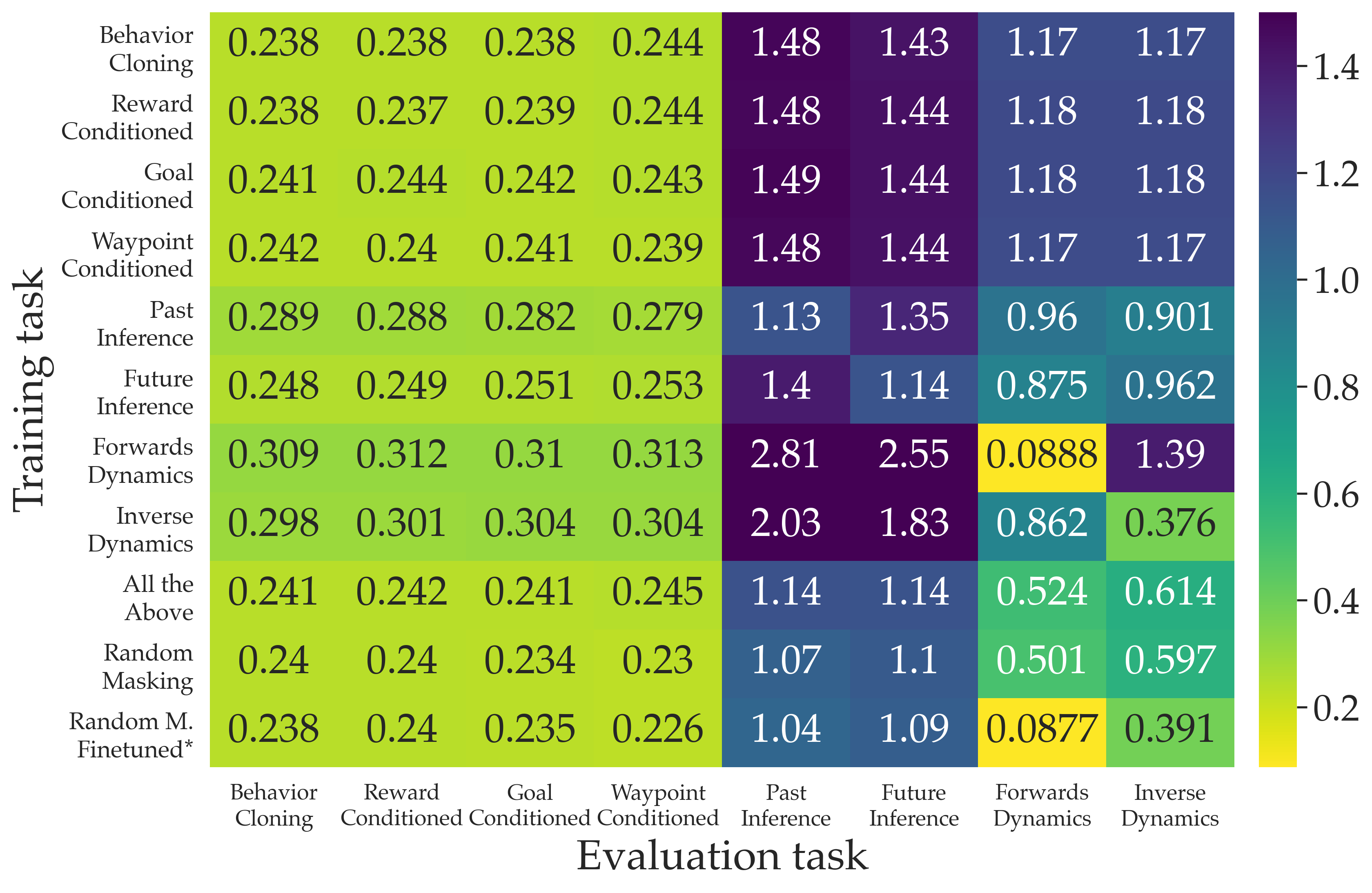}
    \caption{The raw loss values corresponding to \Figref{fig:gridworld-valid-ranks50}.}
    \label{fig:gridworld-valid-raw50}
\end{figure}
\begin{figure}[h!]
    \centering
    \includegraphics[width=0.7\columnwidth]{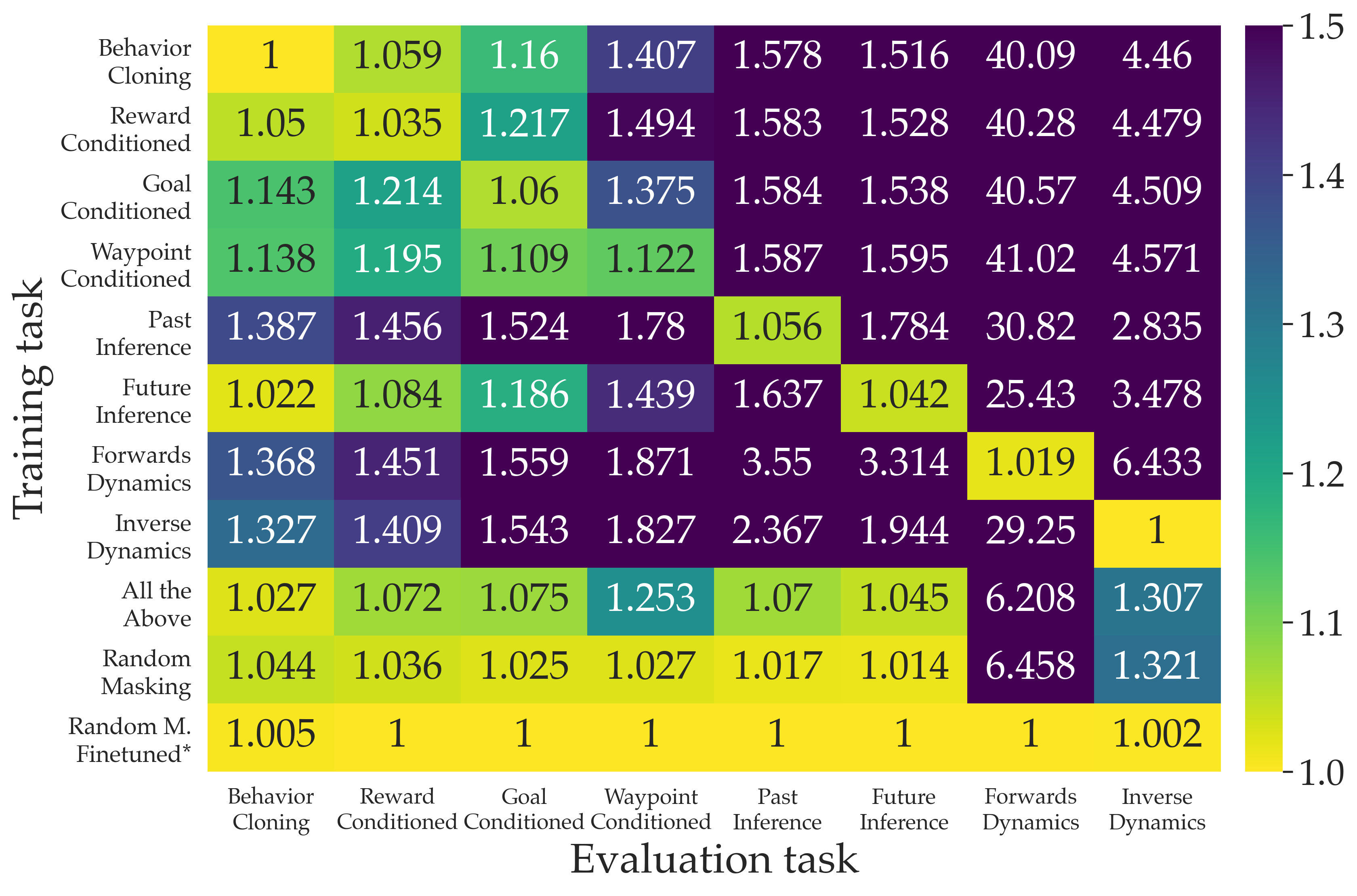}
    \caption{\Figref{fig:gridworld-valid-ranks500} when using a dataset of 1000 trajectories instead of 500. We see that most of the trends remain the same.}
    \label{fig:gridworld-valid-ranks1000}
\end{figure}
\begin{figure}[h!]
    \centering
    \includegraphics[width=0.7\columnwidth]{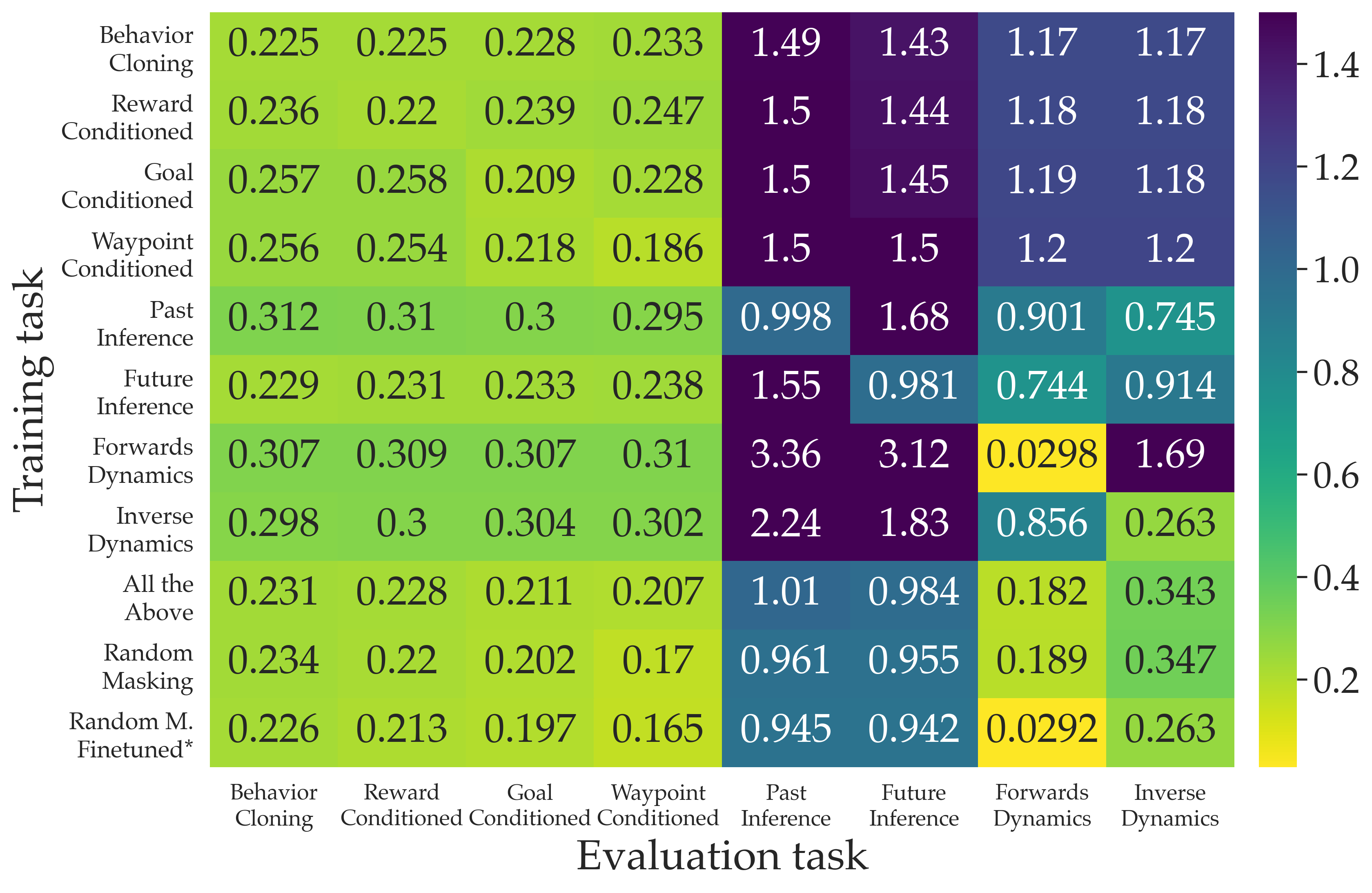}
    \caption{The raw loss values corresponding to \Figref{fig:gridworld-valid-ranks1000}.}
    \label{fig:gridworld-valid-raw1000}
\end{figure}

\end{document}